\documentclass[journal]{IEEEtran}

\usepackage{cite}
\usepackage[pdftex]{graphicx}
\usepackage[cmex10]{amsmath} 
\usepackage{array}

\usepackage{amssymb}  
\usepackage{color}
\usepackage[caption=false]{subfig}
\usepackage{dsfont}
\usepackage{dblfloatfix}

\usepackage{booktabs}
\usepackage[usenames,dvipsnames,svgnames,table]{xcolor}

\newcommand\ie{\textit{i.e.\ }}
\newcommand\eg{\textit{e.g.\ }}

\newcommand{\fig}[1]{Fig.~\ref{fig:#1}}
\newcommand{\secref}[1]{Section~\ref{sec:#1}}

\DeclareMathOperator*{\argmax}{argmax}

\begin{document}
\title{Latent Hierarchical Model for Activity Recognition}
%
%
%

\author{Ninghang~Hu,
        Gwenn~Englebienne,
        Zhongyu~Lou,
        and~Ben~Kr\"{o}se
\thanks{This work was supported by the EU Projects ACCOMPANY (FP7-287624)
and MONARCH (FP7-601033).}
\thanks{The authors are with the Informatics Institute, University of Amsterdam,
The Netherlands (e-mail: n.hu@uva.nl; g.englebienne@uva.nl;
 z.lou@uva.nl; 
 b.j.a.krose@uva.nl).}
\thanks{Manuscript received xx xx, 2014; revised xx xx, 2014.}}

%


\maketitle

\begin{abstract}
We present a novel hierarchical model for human activity
recognition. In contrast to approaches that successively recognize
actions and activities, our approach jointly models actions and
activities in a unified framework, and their labels are simultaneously
predicted. The model is embedded with a latent layer that is able to
capture a richer class of contextual information in both state-state
and observation-state pairs. Although loops are present in the model,
the model has an overall linear-chain structure, where the exact
inference is tractable. Therefore, the model is very efficient in both
inference and learning. The parameters of the graphical model are
learned with a Structured Support Vector Machine (Structured-SVM). A
data-driven approach is used to initialize the latent variables;
therefore, no manual labeling for the latent states is required. The
experimental results from using two benchmark datasets show that our
model outperforms the state-of-the-art approach, and our model is
computationally more efficient.
\end{abstract}


%
\IEEEpeerreviewmaketitle

\section{Introduction\label{sec:introduction}}

\IEEEPARstart{T}{he} use of robots as companions to help people in
their daily life is currently being widely studied.
Numerous studies have focused on providing people with physical
\cite{InterpersonalYoshihiro}, cognitive \cite{TowardsPineau} or
social \cite{LivingKazuyoshi} support. To achieve this, a fundamental
and necessary task is to recognize human activities. For example, to
decide when to offer physical support, a robot needs to recognize that
a person is walking. To decide whether to remind people to continue drinking, a robot needs to recognize past drinking activities. To determine whether a person is lonely, a robot needs to detect interactions between people. In this paper, we propose a hierarchical approach to model human activities.

\begin{figure}[!t]
\centering
\subfloat[]{\includegraphics[width=0.86\linewidth]{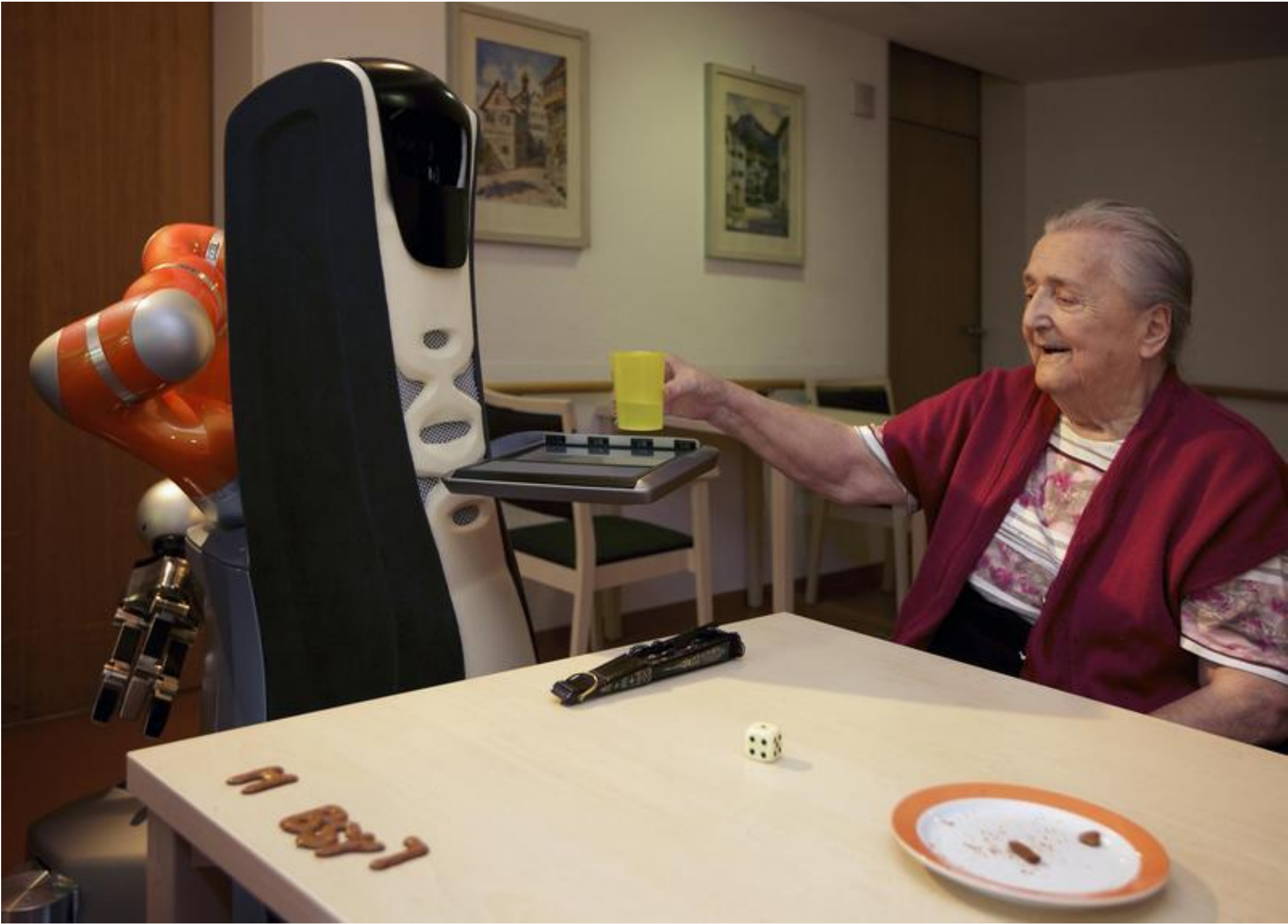}%
\label{fig:cover_girl_color}}
\hfil
\subfloat[]{\includegraphics[width=0.86\linewidth]{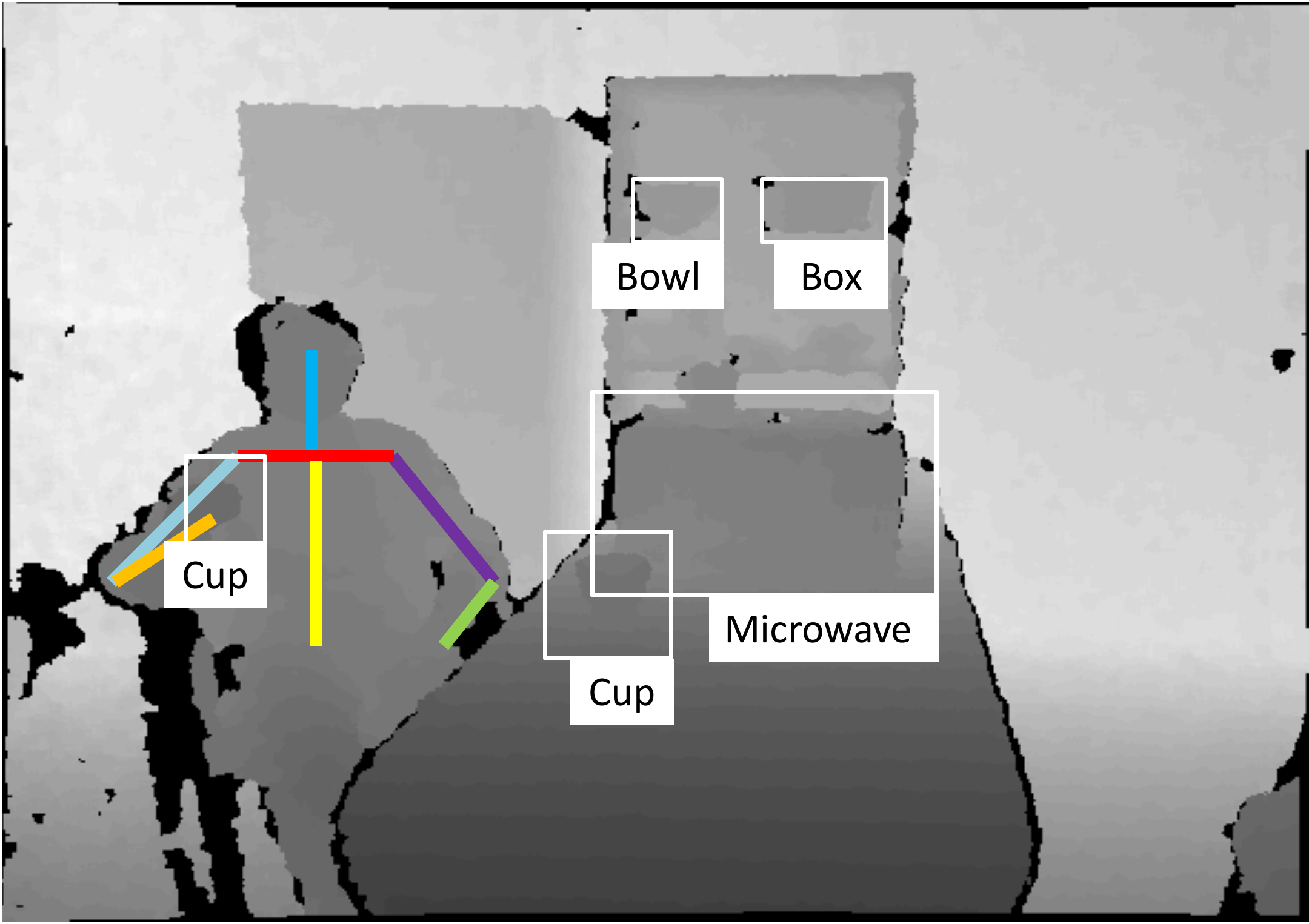}%
\label{fig:cover_girl_depth}}
\caption{An example that shows a robot helping people in an elderly home. \protect\subref{fig:cover_girl_color}
 Care-O-bot 3 offers water to the elderly after detecting that the
 elderly resident has not drunk any water for a long time. \protect\subref{fig:cover_girl_depth}
In this work, an RGB-D sensor is used to recognize human activities. This work is built upon existing methods in object recognition, object localization, and human skeleton tracking. Object and human skeleton information are combined as the input of our model to infer human activities.}
\label{fig:cover_girl}
\end{figure}

Different types of sensors have been applied to the task of activity
recognition \cite{LaraSurvey,ryoo2011human}. Kasteren et
al.\cite{van2010activity} adopt a set of simple sensors, \ie,
pressure, contact, and motion sensors, to recognize daily activities
of people in a smart home. Hu et al.\cite{2013HuPosture} use a
ceiling-mounted color camera to recognize human postures, and the
postures are recognized based on still images. Recently, RGB-D
sensors, such as the Microsoft Kinect and ASUS Xtion Pro, have become
popular in activity recognition because they can capture 2.5D data
using structured light, thereby allowing researchers to extract a rich class of depth features for activity recognition. In this work, we equip a robot with an RGB-D sensor to collect sequences of activity data, from which we extract object locations and human skeleton points, as shown in \fig{cover_girl}. Based on these observations, our task is to estimate \emph{activities} as well as sequences of composing \emph{actions}.

\begin{figure}[t]
\centering
\includegraphics[width=\linewidth]{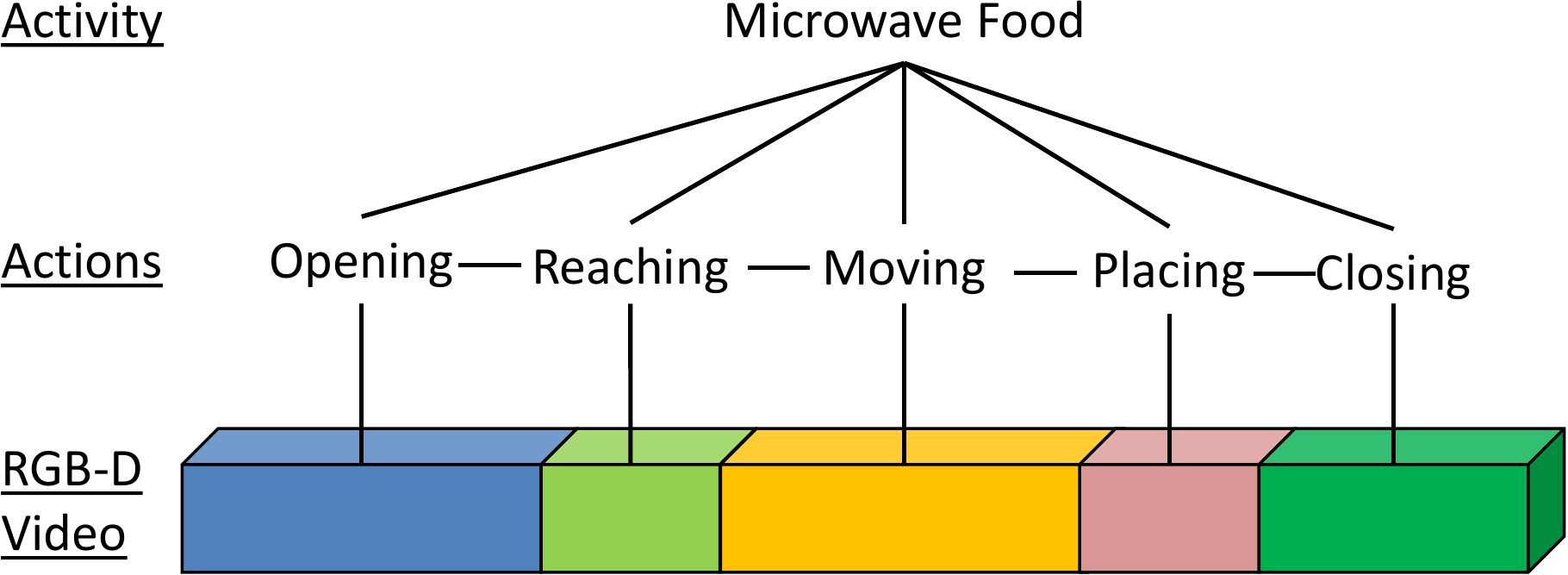}
\caption{An illustration of the activity hierarchy. The input video is represented as a spatial-temporal volume. The bottom layer shows a video that is discretized into multiple temporal segments for modeling, and spatial-temporal features are extracted from each temporal segment. In the middle layer, actions are recognized from the input features with one atomic activity per segment. In the second layer, the activities are described in terms of the sub-level activity sequence. The un-directed links in the graph represent the inter-dependency between layers. Note that the video segments may not have the same length; thus, a segmentation method needs to be applied.}
\label{fig:model}
\end{figure}

We distinguish between \emph{activities} and \emph{actions} as
follows. \emph{Actions} are the atomic movements of a person that
relate to at most one object in the environment, \eg, reaching,
placing, opening, and closing. Most of these actions are completed in
a relatively short period of time. In contrast, \emph{activities} refer to a complete sequence that is composed of different actions. For example, \emph{microwaving food} is an activity that can be decomposed into a number of actions such as \emph{opening} the microwave, \emph{reaching} for  food, \emph{moving} food, \emph{placing} food, and \emph{closing} the microwave. The relation between actions and activities is illustrated in \fig{model}.

The recognition of actions is usually formulated as a sequential prediction problem \cite{Hu2014highlevel} (see \fig{model}). In this approach, the RGB-D video is first divided into smaller video segments so that each segment contains approximately
one action. This can be accomplished either by manual annotation or by
automated temporal segmentation based on motion
features. Spatio-temporal features are extracted for each segment. For
real-world tasks in HRI, it is desirable to recognize activities at a
higher level whereby the activities are usually performed over a
longer duration. The combination of actions and activities forms a
sequential model with a hierarchy (\fig{model}).

Most previous work addresses activity and action recognition as
separate tasks \cite{koppula13IJRR, koppula13icml,Hu2014highlevel},
\ie, the action labels need to be inferred before the
activity labels are predicted. In contrast, in this paper, we jointly
model actions and activities in a unified framework, where the
activity and action labels are learned simultaneously. Our
experimental evaluation demonstrates that this framework is beneficial
when compared to separate recognition. This can be intuitively understood by considering the case of learning actions: the activity label provides additional constraints to the action labels, which can result in a better estimation of the actions, and vice versa. 

\begin{figure}[t]
\centering
\includegraphics[width=0.8\linewidth]{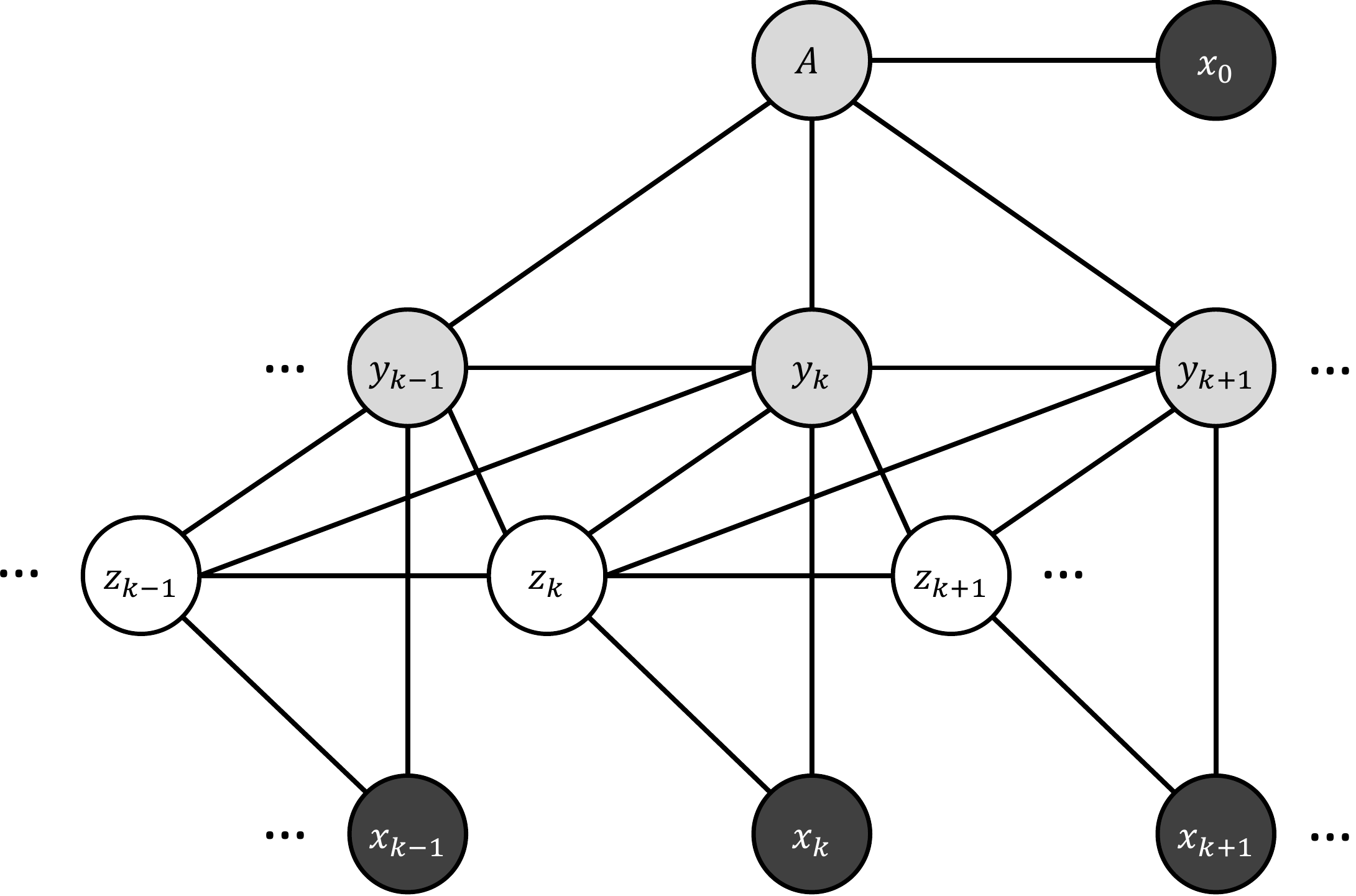}
\caption{The graphical representation of our model. Nodes that
  represent the observations $\boldsymbol{x}$, which are observed both
  in training and testing, are rendered in black. $\boldsymbol{y}$
  refers to action nodes, and $\mathcal{A}$ is the corresponding
  activity label of the sequence. Both are in gray because they are
  only observed during training and not testing. White nodes
  $\boldsymbol{z}$ refer to the latent variables, which are unknown
  either in training or testing. They are used to represent the hidden
  sub-level semantics among consecutive actions. Note that
  $\boldsymbol{x_k}$, $\boldsymbol{y_k}$, $\boldsymbol{z_k}$ are fully
  connected in our model as are the temporal transitions of
  action-latent pairs. Therefore, the model enables a richer
  representation of a activity hierarchy. $\boldsymbol{x}_0$ represents the set of global features.}
\label{fig:graphical_model}
\end{figure} 

\fig{graphical_model} is a graphical representation of our
approach. The proposed model of this paper is based on our previous
work \cite{Hu2014activity}, wherein we recognize the sequence of
actions using Conditional Random Fields (CRFs). The model is augmented
with a layer of latent nodes to enrich the model's expressiveness. For
simplicity, we use \emph{latent variables} to refer to the variables
in the hidden layer, which are unknown both during training and
testing. \emph{Labels}, in contrast, are known during training but are
latent during testing. The latent variables are able to capture such a
difference and are able to model the rich variations of the
actions. One can imagine that the latent variables represent sub-types
of the actions: \eg, for the action \emph{opening}, we are able to
model the difference between \emph{opening a bottle} and \emph{opening
  a door} using latent variables. 

For each temporal segment, we preserve the full connectivity among
observations, latent variables, and action nodes, thus avoiding making
inappropriate conditional independence assumptions. We describe an
efficient method of applying exact inference in our graph, whereby
collapsing the latent states and target states allows our graphical
model to be considered as a linear-chain structure. Applying exact
inference under such a structure is very efficient. We use a
max-margin approach for learning the parameters of the
model. Benefiting from the discriminative framework, our method needs
not model the correlation between the input data, thus providing us
with a natural way of data fusion.

The model was evaluated using the \mbox{RGB-D} data from two different
benchmark datasets\cite{sung2012unstructured,koppula13IJRR}. The
results are compared with a number of the state-of-the-art approaches
\cite{sung2012unstructured,koppula13IJRR,koppula13icml,Hu2014activity,Hu2014highlevel}. The
results show that our model performs better than the state-of-the-art
approaches, and the model is more efficient in terms of inference.

In summary, the contribution of this paper is a novel Hidden CRF model
for jointly predicting activities and their sub-level actions, which
outperforms the state of the art both in terms of predictive performance and in computational cost. Our software is open source and freely accessible at \url{http://ninghanghu.eu/activity_recognition.html}.

In this paper, we address the following research questions:
\begin{itemize}
\item How important is it to add an activity hierarchy to the model?
\item How important is it to add the latent layer to the model?
\item How important is it to joint model actions and activities?
\item How does our model compare with state-of-the-art approaches?
\item How well can the model be generalized to a new problem?
\end{itemize}

The remainder of the paper is organized as follows. We describe the related work in \secref{related_work}. We formalize the model and present the objective function in \secref{model}. The inference and learning algorithms are introduced in \secref{inference} and \secref{learning}. We show the implementation details and the comparison of the results with the state-of-the-art approach in \secref{experiments}.

\section{Related Work \label{sec:related_work}}

The previous works can be categorized into two methodologies. The
first methodology divides the approaches based on the hierarchical
layout of the model, \ie, whether the model contains a single layer or
multiple layers. The second methodology is based on the nature of the
learning method, \ie, whether the method is discriminative or generative.

\subsection{Single-layer Approach and Hierarchical Approach}

Human activity recognition is a key component for HRI, particularly
for the re-ablement of the elderly
\cite{amirabdollahian2013assistive}. Depending on the complexity and
duration of activities, activity recognition approaches can be separated into two categories \cite{aggarwal2011human}: single-layer approaches and hierarchical approaches. Single-layer approaches \cite{laptev,LiuJingen,NiyogiSA,RyooMS,HoaiMinh,Hu2014tracking,
MatikainenPyry,Shi2011,kelley2008understanding} refer to methods that
are able to directly recognize human activities from the data without
defining any activity hierarchy. Usually, these activities are both
simple and short; therefore, no higher level layers are
required. Typical activities in this category include walking,
waiting, falling, jumping and waving. Nevertheless, in the real world,
activities are not always as simple as these basic actions. For
example, the activity of preparing breakfast may consist of multiple
actions such as opening a fridge, getting a salad and making
coffee. Typical hierarchical approaches
\cite{koppula13IJRR,Hu2014activity,Ivanov,Savarese,koppula2013anticipating}
first estimate the sub-level actions, and then, the high-level activity labels are inferred based on the action sequences.

Sung et al. \cite{sung2012unstructured} proposed a hierarchical
maximum entropy Markov model that detects activities from RGB-D
videos. They consider the actions as hidden nodes that are learned
implicitly. Recently, Koppula et al.~\cite{koppula13IJRR} presented an
interesting approach that models both activities and object affordance
as random variables. The object affordance label is defined as the
possible manners in which people can interact with an object, \eg,
reachable, movable, and eatable. These nodes are inter-connected to model object-object and object-human interactions. Nodes are connected across the segments to enable temporal interactions. Given a test video, the model jointly estimates both human activities and object affordance labels using a graph-cut algorithm. After the actions are recognized, the activities are estimated using a multi-class SVM. In this paper, we build a hierarchical approach that jointly estimates actions and activities from the RGB-D videos. The inference algorithm is more efficient compared with graph-cut methods.

\subsection{Generative Models and Discriminative Models}

Many different graphical models, \eg, Hidden Markov Models (HMMs)
\cite{zhu2009human,sung2012unstructured}, Dynamic Bayesian Networks
(DBNs) \cite{ho2009active}, linear-chain CRFs
\cite{vail2007conditional}, loopy CRFs \cite{koppula13IJRR},
Semi-Markov Models \cite{van2010activity}, and Hidden CRFs
\cite{wang2006hidden,wang2009max}, have been applied to the
recognition of human activities. The graphical models can be divided
into two categories: generative models
\cite{zhu2009human,sung2012unstructured} and discriminative models
\cite{koppula13IJRR,van2010activity,2013HuPosture}. The generative
models require making assumptions concerning both the correlation of
data and  how the data are distributed given the activity state. This
is risky because the assumptions may not reflect the true attributes
of the data. The discriminative models, in contrast, only focus on
modeling the posterior probability regardless of how the data are
distributed. The robotic and smart environment scenarios are usually
equipped with a combination of multiple sensors. Some of these sensors
may be highly correlated both in the temporal and spatial domain, \eg,
a pressure sensor on a mattress and a motion sensor above a
bed. In these scenarios, the discriminative models provide a natural way of data fusion for human activity recognition.

The linear-chain Conditional Random Field (CRF) is one of the most
popular discriminative models and has been used for many
applications. Linear-chain CRFs are efficient models because the exact
inference is tractable. However, these models are limited because they cannot capture the intermediate structures within the target states \cite{quattoni2007hidden}. By adding an extra layer of latent variables, the model allows for more flexibility and therefore can be used for modeling more complex data. The names of these models, including Hidden-unit CRF \cite{maaten2011hidden}, Hidden-state CRF \cite{quattoni2007hidden} or Hidden CRF \cite{wang2009max}, are inter-changeable in the literature.

Koppula et al.\ \cite{koppula13IJRR} present a model for the temporal
and spatial interactions between humans and objects in loopy
CRFs. More specifically, they develop a model that has two types of
nodes for representing the action labels of the human and the object
affordance labels of the objects. Human nodes and object nodes within
the same temporal segment are fully connected. Over time, the nodes
are transited to the nodes with the same type. The results show that
by modeling the human-object interaction, their model outperforms the
earlier work in \cite{sung2012unstructured} and
\cite{ni2012order}. The inference in the loopy graph is solved as a
quadratic optimization problem using the graph-cut method
\cite{rother2007optimizing}. Their inference method, however, is less
efficient compared with the exact inference in a linear-chain
structure because the graph-cut method requires multiple iterations
before convergence; more iterations are usually preferred to ensure
that a good solution is obtained.

Another study  \cite{tang2012learning} augments an additional layer of
latent variables to the linear-chain CRFs. They explicitly model the
new latent layer to represent the durations of activities. In contrast
to \cite{koppula13IJRR}, Tang et al.\ \cite{tang2012learning} solve
the inference problem by reforming the graph into a set of cliques so
that the exact inference can be efficiently solved  using dynamic programming. In their model, the latent variables and the observation are assumed to be conditionally independent given the target states.

Our work is different from the previous approaches in terms of both
the utilized graphical model and the efficiency of inference. First,
similar to \cite{tang2012learning}, our model also uses an extra
latent layer. However, instead of explicitly modeling  the latent
variables, we directly learn the latent variables from the
data. Second, we do not make conditional independence assumptions
between the latent variables and the observations. Instead, we add one
extra edge between them to make the local graph fully
connected. Third, although  our graph also presents many loops, as in
\cite{koppula13IJRR}, we are able to transform the cyclic graph into a
linear-chain structure wherein the exact inference is tractable. The
exact inference in our graph only requires two passes of messages
across the linear chain structure, which is substantially more
efficient than the method in \cite{koppula13IJRR}. Finally, we model
the interaction between the human and the objects at the feature level
instead of modeling the object affordance as target states. Therefore,
the parameters are learned and are directly optimized for activity
recognition rather than for making the joint estimation of both object
affordance and human activity. Because we apply a data-driven approach
to initializing the latent variables, hand labeling of the object affordance is not necessary in our model. Our results show that the model outperforms the state-of-the-art approaches on the CAD$120$ dataset \cite{koppula13IJRR}.

\section{Modeling Activity Hierarchy} \label{sec:model}
The graphical model of our proposed system is illustrated in \fig{graphical_model}. Let $\boldsymbol{x}=\{ \boldsymbol{x}_1,\boldsymbol{x}_2,\dots,\boldsymbol{x}_{K}|\boldsymbol{x}_k\in \mathbb{R}^D\}$ be the sequence of observations, where $K$ is the total number of temporal segments in the video. Our goal is to predict the most likely underlying action sequence $\boldsymbol{y}=\{y_1,y_2,\dots,y_{K}|y_k\in \mathcal{Y}\}$ and its corresponding activity label $A\in\mathcal{H}$ based on the observations. We define $\boldsymbol{x}_0$ as the global features that are extracted from $\boldsymbol{x}$.

Each observation $\boldsymbol{x}_k$ is a feature vector extracted from
the segment $k$. The form of $\boldsymbol{x}_k$ is quite
flexible. $\boldsymbol{x}_k$ can be collections of data from different
sources, \eg, simple sensor readings, human locations, human poses,
and object locations. Some of these observations may be highly
correlated with each other, \eg, wearable accelerate meters and motion
sensors would be highly correlated. Because of the discriminative nature of our model, we do not need to model such correlation among the observations. 

We define $\boldsymbol{z}=\{z_1,z_2,\dots,z_{K}|z_k\in \mathcal{Z}\}$
to be the latent variables in the model. The latent variables, which
are implicitly learned from the data, can be considered as modeling
the sub-level semantics of the actions. For clarity, one could imagine that $y_k=1$ refers to the action \emph{opening}. Then, the joint $(y_k=1,z_k=1)$ can describe \emph{opening a microwave}, and $(y_k=1,z_k=2)$ can describe \emph{opening a bottle}. Note that these two sub-types of \emph{opening} actions differ greatly in the observed videos. However, the latent variables allow us to capture large variations in the same action.

Next, we will formulate our model in terms of these defined
variables. For simplicity, we assume that there are in total $N_y$ actions and $N_a$ activities to be recognized, and let us define $N_z$ as the cardinality of the latent variable.

\subsection{Potential Function}
Our model contains five types of potentials that together form the potential function. 

The first potential measures the score of making an observation $\boldsymbol{x}_k$ with a joint-state assignment $(z_k,y_k$). We define $\Phi(\boldsymbol{x}_k)$ to be the function that maps the input data into the feature space. $\boldsymbol{w}$ is a matrix that contains model parameters.
\begin{equation}
\psi_1(y_k,z_k,\boldsymbol{x}_k;\boldsymbol{w}_1)= \boldsymbol{w}_1(y_k,z_k) \cdot \Phi(\boldsymbol{x}_k)
\label{eq:psi1}
\end{equation}
where $\boldsymbol{w}_1\in\mathbb{R}^{\mathcal{Y}\times\mathcal{Z}\times D}$ and $\boldsymbol{w}_1(y_k,z_k)$ is the concatenation of the parameters that corresponds to $y_k$ and $z_k$.

This potential models the full connectivity among $y_k$, $z_k$ and
$\boldsymbol{x}_k$ and avoids making any conditional independence
assumptions. It is more accurate to have such a structure because
$z_k$ and $\boldsymbol{x}_k$ may not be conditionally independent over
a given $y_k$ in many cases. Let us consider the aforementioned
example. Knowing that the action is \emph{opening}, whether the latent state refers to \emph{opening} a microwave or \emph{opening} a bottle depends on how the \emph{opening} action is performed in the observed video, \ie, the latent state and the observation are inter-dependent given the action label.

The second potential measures the score of coupling $y_k$ with
$z_k$. The score can be considered as either the bias entry of \eqref{eq:psi1} or the prior of seeing the joint state $(y_k,z_k)$.
\begin{equation}
\psi_2(y_k,z_k;\boldsymbol{w}_2)= \boldsymbol{w}_2(y_k,z_k)
\end{equation}
where $\boldsymbol{w}_2$ represents the parameter of the second
potential with $\boldsymbol{w}_2\in\mathbb{R}^{\mathcal{Y}\times\mathcal{Z}}$.

The third potential characterizes the transition score from the joint
state $(y_{k-1},z_{k-1})$ to $(y_{k},z_{k})$. Comparing with the
normal transition potentials \cite{wang2009max}, our model leverages
the latent variable $z_k$ for modeling richer contextual information
over consecutive temporal segments. Our model not only contain the transition between the action states but also captures the sub-level context using the latent variables.
\begin{equation}
\psi_3(y_{k-1},z_{k-1},y_k,z_k;\boldsymbol{w}_3)=  \boldsymbol{w}_3(y_{k-1},z_{k-1},y_k,z_k)
\end{equation}
where the potential is parameterized by $\boldsymbol{w}_3\in \mathbb{R}^{\mathcal{Y}\times\mathcal{Z}\times\mathcal{Y}\times\mathcal{Z}}$.

The fourth potential models the compatibility among consecutive action pairs and the activity. 
\begin{equation}
\psi_4(y_{k-1},y_k,A;\boldsymbol{w}_4)=  \boldsymbol{w}_4(y_{k-1},y_k,A)
\end{equation}
where $\boldsymbol{w}_4 \in
\mathbb{R}^{\mathcal{Y}\times\mathcal{Y}\times\mathcal{H}}$ and
$\boldsymbol{w}_4(y_{k-1},y_k,A)$ is a scalar that reflects the
compatibility between the transition of an action and the activity label.

The last potential models the compatibility between the activity label $A$ and the global features $\boldsymbol{x}_0$.
\begin{equation}
\psi_5(A,\boldsymbol{x}_0;\boldsymbol{w}_5)= \boldsymbol{w}_5(A) \cdot \Phi(\boldsymbol{x}_0)
\end{equation}
where the parameters $\boldsymbol{w}_5 \in \mathbb{R}^{\mathcal{H}}$ can be interpreted as a global filter that favors certain combinations of $\boldsymbol{x}_0$ and $A$.

Summing all potentials over the entire sequence, we can write the final potential function of our model as follows:
\begin{align}\label{eq:objective_function}
&F(A,\boldsymbol{y},\boldsymbol{z},\boldsymbol{x};\boldsymbol{w}) =
\sum_{k=1}^K \boldsymbol{w}_1(y_k,z_k) \cdot \Phi(\boldsymbol{x}_k) + \sum_{k=1}^K\boldsymbol{w}_2(y_k,z_k) \nonumber\\
&+ \sum_{k=2}^K \boldsymbol{w}_3(y_{k-1},z_{k-1},y_k,z_k) + \sum_{k=2}^K\boldsymbol{w}_4(y_{k-1},y_k,A)\nonumber \\
&+ \boldsymbol{w}_5(A) \cdot \Phi(\boldsymbol{x}_0)
\end{align}

The potential function evaluates the matching score between the joint
states $(A,\boldsymbol{y},\boldsymbol{z})$ and the input
$(\boldsymbol{x}$, $\boldsymbol{x}_0)$. The score equals the
un-normalized joint probability in the log space. The objective
function can be rewritten in a more general linear form $F(\boldsymbol{y},\boldsymbol{z},\boldsymbol{x};\boldsymbol{w}) = \boldsymbol{w}\cdot \Psi(\boldsymbol{y},\boldsymbol{z},\boldsymbol{x})$. Therefore, the model is in the class of log-linear models.

Note that it is not necessary to explicitly model the latent
variables; rather, the latent variables can be automatically learned
from the training data. Theoretically, the latent variables can
represent any form of data, \eg, time duration and action primitives,
as long as the data can be used to facilitate the task. The optimization of the latent model, however, may converge to a local minimum. The initialization of the random variables is therefore of great importance. We compare three initialization strategies in this paper. Details of the latent variable initialization will be discussed in \secref{init_latent}.

One may notice that our graphical model has many loops, which in
general makes the exact inference intractable. Because our graph
complies with the semi-Markov property, we will now show how we
benefit from such a structure to obtain efficient inference and learning.

\section{Inference \label{sec:inference}}
Given the graph and the parameters, inference is used to find the most likely joint state $(A,\boldsymbol{y},\boldsymbol{z})$ that maximizes the objective function.
\begin{equation}
(A^*,\boldsymbol{y}^*,\boldsymbol{z}^*)=\argmax_{(A,\boldsymbol{y},\boldsymbol{z}) \in \mathcal{H}\times\mathcal{Y}\times\mathcal{Z}} F(A,\boldsymbol{y},\boldsymbol{z},\boldsymbol{x};\boldsymbol{w})
\label{eq:inferences}
\end{equation}

Generally, solving \eqref{eq:inferences} is an NP-hard problem that
requires the evaluation of the objective function over an exponential
number of state sequences. Exact inference is preferable because it is
guaranteed to find the global optimum. However, the exact inference
usually can only be applied efficiently when the graph is acyclic. In
contrast, approximate inference is more suitable for loopy graphs but
may take longer to converge and is likely to obtain a local
optimum. Although our graph contains loops, we can transform the graph
into a linear-chain structure, in which the exact inference becomes
tractable. If we collapse the latent variable $z_k$ with $y_k$ and $A$
into a single factor, the edges among $z_k$, $y_k$ and $A$ become the
internal factor of the new node, and the transition edges collapse
into a single transition edge. This results in a typical linear-chain
CRF, where the cardinality of the new nodes is $N_y\times N_z\times
N_A$. In the linear-chain CRF, the exact inference can be efficiently performed using dynamic programming \cite{bellman1986dynamic}.

Using the chain property, we can write the following recursion procedure for computing the maximal score over all possible assignments of $\boldsymbol{y}$, $\boldsymbol{z}$ and A.
\begin{align} \label{eq:max_sum}
V_k(A,y_k,z_k) = &\boldsymbol{w}_1(y_k,z_k) \cdot \Phi(\boldsymbol{x}_k) + \boldsymbol{w}_2(y_k,z_k) \nonumber\\
& +\max_{(y_{k-1},z_{k-1})\in \mathcal{Y}\times \mathcal{Z}} \{ \boldsymbol{w}_3(y_{k-1},z_{k-1},y_k,z_k) \nonumber\\
&+\boldsymbol{w}_4(y_{k-1},y_k,A)+ V_{k-1}(A,y_{k-1},z_{k-1}) \}
\end{align}

The above function is evaluated iteratively across the entire sequence. For each iteration, we record the joint state $(y_{k-1},z_{k-1})$ that contributes to the max. When the last segment is computed, the optimal assignment of segment $K$ can be computed as
\begin{equation}
A^*,y_K^*,z_K^* = \argmax_{A,y_K,z_K}V_K(A,y_K,z_K)+ \boldsymbol{w}_5(A) \cdot \Phi(\boldsymbol{x}_0)
\end{equation}

Knowing the optimal assignment at $K$, we can track back the best assignment in the previous time step $K-1$. The process continues until all $\boldsymbol y^*$ and $\boldsymbol z^*$ have been assigned, \ie, the inference problem in \eqref{eq:inferences} is solved.

Computing \eqref{eq:max_sum} once involves $O(N_yN_z)$
computations. In total, \eqref{eq:max_sum} needs to be evaluated for
all possible assignments of $(y_k,z_k,N_A)$; thus, it is computed $N_yN_z$ times. The total computational cost is, therefore, $O(N_y^2\,N_z^2\,N_AK)$. Such computation is manageable when $N_yN_z$ is not very large, which is usually the case for the tasks of activity recognition.

Next, we show how we can learn the parameters using the max-margin approach.

\section{Learning \label{sec:learning}}
We use the max-margin approach for learning the parameters in our graphical model. Given a set of N training examples, $\langle\boldsymbol{x}^{(n)},\boldsymbol{y}^{(n)},A^{(n)}\rangle$ $(n=1, 2, \cdots, N)$, we would like to learn the model parameters $\boldsymbol{w}$ that can produce the activity label $A$ and action labels $\boldsymbol{y}$ given a new test input $\boldsymbol{x}$. Note that both activities and action labels are observed during training. The latent variables $\boldsymbol{z}$ are unobserved and will be automatically inferred from the training process. 

The goal of learning is to find the optimal model parameters $\boldsymbol{w}$ that minimize the objective function. A regularization term is used to avoid over-fitting.
\begin{equation}
\min_{\boldsymbol{w}} \left\{\frac{1}{2} \|\boldsymbol{w}\|^2 + C \sum_{i=1}^N \Delta(\boldsymbol{y}^{(i)},\widehat{\boldsymbol{y}},A^{(i)},\widehat{A}) \right\}
\label{eq:loss_basic}
\end{equation}
where $C$ is a normalization constant that is used to provide a
balance between the model complexity and fitting rate. 

The loss function
$\Delta(\boldsymbol{y}^{(i)},\widehat{\boldsymbol{y}},A^{(i)},\widehat{A})$
measures the cost of making incorrect
predictions. $\widehat{\boldsymbol{y}}$ and $\widehat{A}$ are the most
likely action and activity labels that are computed from
\eqref{eq:inferences}. The loss function in \eqref{eq:loss_delta}
returns zero when the prediction is exactly the same as the ground
truth; otherwise, it counts the number of disagreed elements. 
\begin{equation}
\Delta(\boldsymbol{y}^{(i)},\widehat{\boldsymbol{y}},A^{(i)},\widehat{A})=
\lambda\, \mathds{1}(A^{(i)}=\widehat{A})
+ \frac{1}{K}\sum_{k=1}^{K} \mathds{1}(y_k^{(i)}=\widehat{y}_k)
\label{eq:loss_delta}
\end{equation}
where $\mathds{1}(\cdot)$ is an indicator function and $0\leq \lambda
\leq 1$ is a scalar weight that balances between the two loss terms. 

This object function can be viewed as a generalized form of our
previous work \cite{Hu2014activity}, where we recognize only the
sequence of actions. This is can be performed by simply setting
$\lambda$ to $0$ and leaving the graphical structure unchanged. The
learning framework then only tracks the incorrectly predicted actions, regardless of the activities.

Directly optimizing \eqref{eq:loss_basic} is not possible because the loss function involves computing the $\argmax$ in \eqref{eq:inferences}. Following \cite{tsochantaridis2005large} and \cite{yu2009learning}, we substitute the loss function in \eqref{eq:loss_basic} by the margin rescaling surrogate, which serves as an upper-bound of the loss function.

\begin{align}
&\min_{\boldsymbol{w}} \{\frac{1}{2} \|\boldsymbol{w}\|^2 + C \sum_{i=1}^n \max_{A,\boldsymbol{y},\boldsymbol{z}} [\Delta(\boldsymbol{y}^{(i)},\boldsymbol{y},A^{(i)},A) + \nonumber\\
&F(\boldsymbol{x}^{(i)},\boldsymbol{y},A,\boldsymbol{z},\boldsymbol{w})]
- C \sum_{i=1}^n \max_{\boldsymbol{z}} F(\boldsymbol{x}^{(i)},\boldsymbol{y}^{(i)},A^{(i)},\boldsymbol{z},\boldsymbol{w}) \}
\label{eq:structured_svm}
\end{align}

The second term in \eqref{eq:structured_svm} can be solved using the augmented inference, \ie, by plugging in the loss function as an extra factor in the graph, the term can be solved in the same way as the inference problem using \eqref{eq:inferences}. Similarly, the third term of \eqref{eq:structured_svm} can be solved by adding $\boldsymbol{y}^{(i)}$ and $A^{(i)}$ as the evidence into the graph and then applying inference using \eqref{eq:inferences}. Because the exact inference is tractable in our graphical model, both of the terms can be computed very efficiently.

Note that \eqref{eq:structured_svm} is the summation of a convex and
concave function. This can be solved with the Concave-Convex Procedure
(CCCP) \cite{yuille2002concave}. By substituting the concave function
with its tangent hyperplane function, which serves as an upper bound
of the concave function, the concave term is transformed into a linear function. Thus, \eqref{eq:structured_svm} becomes convex again.

We can rewrite \eqref{eq:structured_svm} in the form of minimizing a function subject to a set of constraints by adding slack variables
\begin{align}\label{eq:structureSVM}
\min_{\boldsymbol{w},\boldsymbol{\xi}} \{\frac{1}{2} \|\boldsymbol{w}\|^2 + &C \sum_{i=1}^n \xi_i \}\\ \nonumber
s.t. ~\forall i \in \{1,2,\dots,n\}, \forall y \in \mathcal{Y}&\nonumber\\
F(\boldsymbol{x}^{(i)},\boldsymbol{y}^{(i)},A^{(i)},\boldsymbol{z}^*,\boldsymbol{w}) - &F(\boldsymbol{x}^{(i)},\boldsymbol{y},A,\boldsymbol{z},\boldsymbol{w}) \nonumber\\ 
&\ge \Delta(\boldsymbol{y}^{(i)},\boldsymbol{y},A^{(i)},A) - \xi_i\nonumber
\end{align}
where $\boldsymbol{z}^*$ is the most likely latent states that are inferred given the training data.

Note that there are an exponential number of constraints in \eqref{eq:structureSVM}. This can be solved using the cutting-plane method \cite{kelley1960cutting}.

Another intuitive method of understanding the CCCP algorithm is to
consider the algorithm as one that solves the learning problem with incomplete data using Expectation-Maximization (EM) \cite{EMalgorithm}. In our training data, the latent variables are unknown. We can start by initializing the latent variables. Once we have the latent variables, the data become complete. Then, we can use the standard Structured-SVM to learn the model parameters (M-step). Subsequently, we can update the latent states again using the parameters that are learned (E-step). The iteration continues until convergence. 

The CCCP algorithm decreases the objective function in each
iteration. However, the algorithm cannot guarantee a global
optimum. To avoid being trapped in a local minimum, we present three different initialization strategies, and details will be presented in \secref{init_latent}.

Note that the inference algorithm is extensively used in learning. Because we are able to compute the exact inference by transforming the loopy graph into a linear-chain graph, our learning algorithm is much faster and more accurate compared with the other approaches with approximate inference.

\section{Experiments and Results \label{sec:experiments}}
We implemented the proposed model, denoted as \emph{full model}, along
with its three variations. Specifically, the first model recognizes
only low-level actions, the second model recognizes high-level
activities, and the third model recognizes activities based on
actions. All models were evaluated on two different datasets. The
results from the different models were compared to gain insight into
our research questions (in \secref{introduction}). The \emph{full
  model} is also shown to  outperform the state-of-the-art methods.

\subsection{Datasets} \label{sec:dataset}
The methods were evaluated on two benchmark datasets, \ie,
\mbox{CAD-60} \cite{sung2012unstructured} and \mbox{CAD-120}
\cite{koppula13IJRR}. Both of the datasets contain sequences of color
and depth images that were collected by a \mbox{RGB-D} sensor. Skeleton joints of the person are obtained using OpenNI\footnote{http://structure.io/openni}. 

\begin{figure*}[!t]
\captionsetup[subfigure]{labelformat=empty}
\centering
\subfloat[Arranging Objects]{\includegraphics[width=0.19\linewidth]{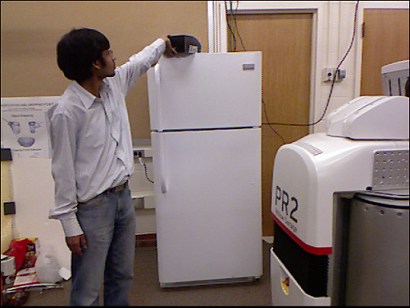}
\label{fig:arranging_objects}}
\subfloat[Cleaning Objects]{\includegraphics[width=0.19\linewidth]{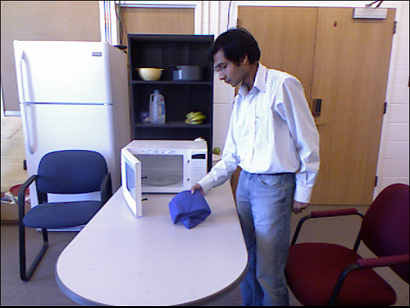}
\label{fig:cleaning_objects}}
\subfloat[Having Meal]{\includegraphics[width=0.19\linewidth]{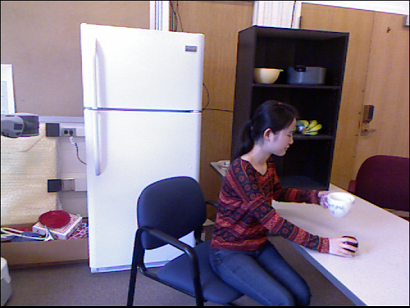}%
\label{fig:having_meal}}
\subfloat[Making Cereal]{\includegraphics[width=0.19\linewidth]{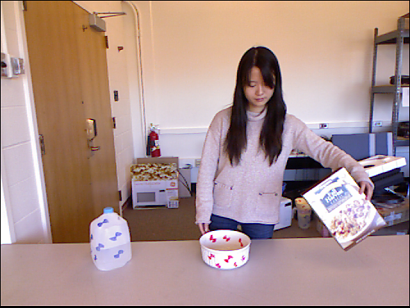}
\label{fig:making_cereal}}
\subfloat[Microwaving Food]{\includegraphics[width=0.19\linewidth]{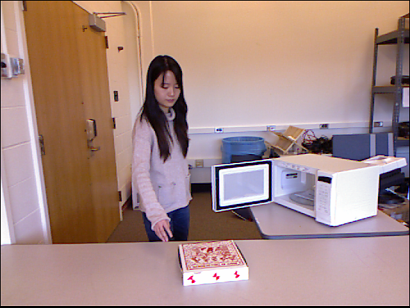}
\label{fig:microwaving_food}}
\\
\subfloat[Picking Objects]{\includegraphics[width=0.19\linewidth]{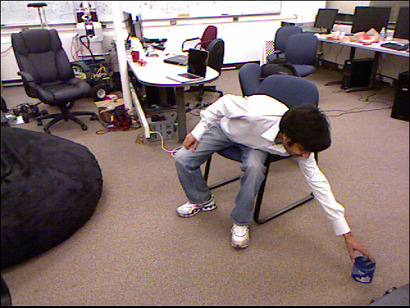}
\label{fig:picking_objects}}
\subfloat[Stacking Objects]{\includegraphics[width=0.19\linewidth]{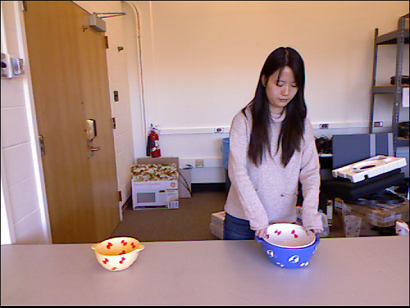}
\label{fig:stacking_objects}}
\subfloat[Taking Food]{\includegraphics[width=0.19\linewidth]{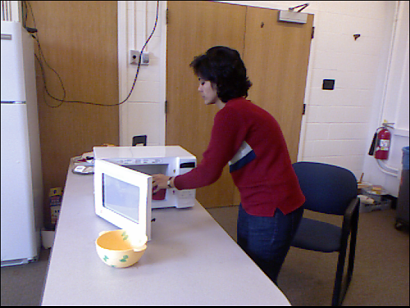}%
\label{fig:taking_food}}
\subfloat[Taking Medicine]{\includegraphics[width=0.19\linewidth]{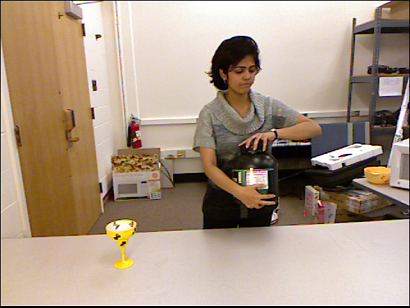}
\label{fig:taking_medicine}}
\subfloat[Unstacking Objects]{\includegraphics[width=0.19\linewidth]{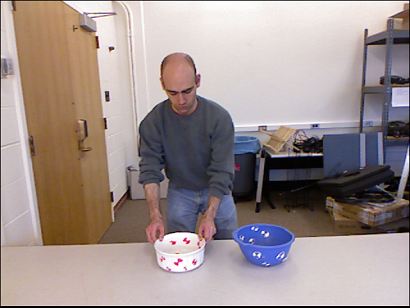}
\label{fig:unstacking_objects}}
\caption{Sample images from the \mbox{CAD-120} dataset. The images illustrate 10 different activities.}
\label{fig:cad120_samples}
\end{figure*}

The two datasets are quite different from each other; therefore, they
can be used to test the generalizability of our methods.
The \mbox{CAD-60} dataset consists of 12 human \emph{action} labels
and no \emph{activity} labels. The \emph{actions} include
\emph{rinsing mouth}, \emph{brushing teeth}, \emph{wearing contact
  lens}, \emph{talking on the phone}, \emph{drinking water},
\emph{opening pill container}, \emph{cooking (chopping)},
\emph{cooking (stirring)}, \emph{talking on couch}, \emph{relaxing on
  couch}, \emph{writing on white board}, and \emph{working on
  computer}. These actions are performed by 4 different subjects in 5
different \emph{environments}, \ie, a kitchen, a bedroom, a bathroom,
a living room, and an office. In total, the dataset includes
approximately 60 videos, and each video contains one action label. In
contrast, the \mbox{CAD-120} dataset \cite{koppula13IJRR} contains 126
RGB-D videos, and each video contains one \emph{activity} and a
sequence of \emph{actions}. There are in total 10 activities defined
in the dataset, including \emph{making cereal}, \emph{taking
  medicine}, \emph{stacking objects}, \emph{unstacking objects},
\emph{microwaving food}, \emph{picking up objects}, \emph{cleaning
  objects}, \emph{taking food}, \emph{arranging objects}, and
\emph{having a meal}. \fig{cad120_samples} shows various sample images
of these activities. In addition, the dataset also consists of 10
sub-level actions, \ie, \emph{reaching, moving, pouring, eating,
  drinking, opening, placing, closing, scrubbing}, and
\emph{null}. The objects in \mbox{CAD-120} are automatically detected
as in \cite{koppula13IJRR}, and the locations of the objects are also provided by the dataset.

The two datasets are very challenging in the following aspects. a) The
activities in the dataset are performed by four different actors. The
actors behave quite differently, \eg, in terms of being left or right
handed, being viewed from a front view or side view, and sitting or
standing. b) There is a large variation even for the same action, \eg,
the action \emph{opening} can refer to opening a bottle or opening the
microwave. Although both of actions have the same label, they appear
significantly different from each other in the video. c) Partial or
full occlusion is also a very challenging aspect for this dataset. \eg
in certain videos, the actors' legs are completely occluded by the table, and objects are frequently occluded by the other objects. This makes it difficult to obtain accurate object locations as well as body skeletons; therefore, the generated data are noisy.

A number of recent approaches
\cite{koppula13IJRR,koppula13icml,Hu2014activity,Hu2014highlevel,sung2012unstructured}
have been evaluated on these two datasets; therefore, the results can
be directly compared. To ensure a fair comparison, the same input features are extracted following \cite{koppula13IJRR}. Specifically, we have object features $\phi_o(\boldsymbol{x}_k) \in \mathbb{R}^{180}$, object-object interaction features $\phi_{oo}(\boldsymbol{x}_k) \in \mathbb{R}^{200}$, object-subject relation features $\phi_{oa}(\boldsymbol{x}_k) \in \mathbb{R}^{400}$, and the temporal object and subject features $\phi_t(\boldsymbol{x}_k) \in \mathbb{R}^{200}$. For \mbox{CAD-120}, we extract the complete set of features from the object locations, which are provided by the dataset. For \mbox{CAD-60} dataset, only skeletal features are extracted because there is no object information. These features are concatenated into a single feature vector, which is considered as the observation of one action segment, \ie, $\Phi(\boldsymbol{x}_k)$.

\subsection{Implemented Models}

In this section, we describe the three baseline models in detail, followed by the introduction of the \emph{full model}. Note that all of the models were first evaluated on the \mbox{CAD-120} dataset. To test the generalizability of the model, the same experiments are repeated using the \mbox{CAD-60} dataset. Note that the CAD-60 dataset contains no \emph{activity} labels, but it has additional labels to indicate the environments. In our experiments, we treat these additional labels as if they were \emph{activities}; thus, the model structure is left unchanged compared with the experiments on CAD-120. The only difference is that we jointly model the actions with the environments instead of the activities.

\subsubsection{Recognize Only Low-level Actions}\label{sec:init_latent}
The first model is adopted from our previous work
\cite{Hu2014activity}, which predicts action labels based on the video
sequence. This model is a single-layer approach that only contains the
low-level layer of nodes. By setting the weight $\lambda$ to zero, the
model focuses only on predicting correct action labels regardless of
the activity label; therefore, the model can be considered as a special case of the full model. The parameters of this model are learned with the Structured-SVM framework \cite{tsochantaridis2005large}. We use the margin rescaling surrogate as the loss. For optimization, we use the 1-slack algorithm (primal), as described in \cite{joachims2009cutting}.

To initialize the latent states on \mbox{CAD-120}, we adopt three
different initialization strategies. a) Random initialization. The
latent states are randomly selected. b) A data-driven approach. We
apply clustering on the input data $\boldsymbol{x}$. The number of
clusters is set to be the same as the number of latent states. We run
\mbox{K-means} for 10 times. Then, we choose the best clustering
results based on minimal within-cluster distances. The labels of the clusters are assigned as the initial latent states. c) Initialized by Object affordance. The object affordance labels are provided by the \mbox{CAD-120} dataset, which are used for training in \cite{koppula13IJRR}. We apply the \mbox{K-means} clustering upon the affordance labels. As the affordance labels are categorical, we use \mbox{1-of-N} encoding to transform the affordance labels into binary values for clustering. The \mbox{CAD-60} dataset do not contain affordance labels, therefore the latent variables are initialized only with the data-driven approach.

\subsubsection{Recognize Only High-level Activities}
The second model contains only a single layer for recognizing
activities, \ie, we disregard the layer of actions; instead, we learn a direct mapping from video features to activity labels. Similar to the first model, the parameters are learned with the Structured-SVM, but the model contains no transition. 

\subsubsection{Recognize Activities Based on Action Sequences}
This approach is built upon the first baseline. Based on the inferred action labels, we learn a model to classify activities. We extract unigram and bigram features based on the action sequence as well as the occlusion features. The model parameters are estimated with a variation of multi-class SVM, where the latent layer is augmented in the model. In this approach, the actions and activities are recognized in succession.

\subsubsection{Joint Estimation of Activity and Actions using Hierarchical Approach (full model)}
This approach refers to the proposed model of the paper. Instead of
successively recognizing actions and activities, our model uses a hierarchical framework to make joint predictions over both activity and action labels.

We compare two different segmentation methods to the videos in the
CAD-120 dataset. In the first method, we use the ground truth
segmentation, which is manually annotated. For the second
segmentation, we apply a motion-based approach, \ie, we extract the
spatial-temporal features for all the frames, and similar frames are
grouped together  using a graph-based approach to form segments. For CAD-60, we apply uniform segmentation as in \cite{koppula13IJRR} to enable a fair comparison with other methods.

The above methods were evaluated on both the \mbox{\mbox{CAD-120}} and \mbox{\mbox{CAD-60}} datasets. Because the two datasets are quite different from each other, they can be used to test how the results can be generalized to new data. The performance of these methods on both datasets is reported in \secref{results}.

\subsection{Evaluation Criteria}\label{sec:evaluation}
Our model was evaluated with 4-fold cross-validation. The folds are
split based on the 4 subjects. To choose the hyper-parameters, \ie,
the number of latent states and segmentation methods, we used two
subjects for training, one subject for validation and one subject for
testing. Once the optimal hyper-parameters are chosen, the performance
of the model during testing is measured by another cross-validation
process, \ie, training using videos of 3 persons and testing on a
\textit{new person}. Each cross-validation is performed 3 times. To
observe the generalization of our model across different data sets,
the results are averaged across the folds. In this paper, the accuracy
(classification rate), precision, recall and F-score are reported to
enable a comparison of the results. In the \mbox{CAD-120} dataset,
more than half of the instances are \emph{reaching} and
\emph{moving}. Therefore, we consider precision and recall to be
relatively better evaluation criteria than accuracy because they remain meaningful despite class imbalance.

\subsection{Results and Analysis}\label{sec:results}
\begin{table*}
\renewcommand{\arraystretch}{1.3}
\caption{Performance of Activity and Action Recognition during testing
  on the \mbox{\mbox{CAD-120}} Dataset. The results are reported in
  terms of Accuracy, Precision, Recall and F-Score. The standard error is also reported.}
\label{tab:results_cad120}
\centering
\begin{tabular}{lcccccccc}
\toprule
\multicolumn{9}{c}{Ground Truth Segmentation}\\
\midrule
Methods & \multicolumn{4}{c}{Action} & \multicolumn{4}{c}{Activity} \\
\cmidrule(r){2-5}\cmidrule(r){6-9}
 & Accuracy       & Precision      & Recall       & F1-Score & Accuracy       & Precision      & Recall   & F-Score\\
\midrule
Single layer & - & - & - & - & $74.2 \pm 5.1$ & $78.5 \pm 4.7$ & $73.3 \pm 5.1$ & $75.8 \pm 4.9$ \\
Koppula et al. \cite{koppula13IJRR} & $86.0 \pm 0.9$ & $84.2 \pm 1.3$ & $76.9 \pm 2.6$ & $80.4 \pm 1.7$ & $84.7 \pm 2.4$  & $85.3 \pm 2.0$ & $84.2 \pm 2.5$ & $84.7 \pm 2.2$\\
Koppula et al. \cite{koppula13icml} & $89.3 \pm 0.9$ & $87.9 \pm 1.8$ & $84.9 \pm 1.5$ & $86.4 \pm 1.6$ & $93.5 \pm 3.0$  & $95.0 \pm 2.3$ & $93.3 \pm 3.1$ & $94.1 \pm 2.6$\\
Hu et al. \cite{Hu2014highlevel}\cite{Hu2014activity} & $87.0 \pm 0.9$ & $89.2 \pm 2.3$ & $83.1 \pm 1.2$ & $85.5 \pm 1.6$ & $90.0 \pm 2.9$  & $92.8 \pm 2.3$ & $89.7 \pm 3.0$ & $91.2 \pm 2.5$\\
Our Model (no latent) & $87.2 \pm 0.8$ & $87.4 \pm 1.5$ & $85.0 \pm 1.4$ & $86.2 \pm 1.4$ & $87.9 \pm 1.5$ & $91.9 \pm 0.7$ & $87.5 \pm 1.6$ & $89.7 \pm 1.0$\\ 
Our Model (full) & $89.7 \pm 0.6$ & $90.2 \pm 0.7$ & $88.2 \pm 0.6$ & $89.2 \pm 0.6$ & $93.6 \pm 2.7$ & $95.2 \pm 2.0$ & $93.3 \pm 2.8$ & $94.2 \pm 2.3$\\
\midrule				
\multicolumn{9}{c}{Motion-based Segmentation}\\
\midrule
Methods & \multicolumn{4}{c}{Action} & \multicolumn{4}{c}{Activity} \\
\cmidrule(r){2-5}\cmidrule(r){6-9}
& Accuracy       & Precision      & Recall       & F-Score & Accuracy       & Precision      & Recall & F-Score   \\
\midrule
Single layer & - & - & - & - & $75.0 \pm 5.3$ & $79.0 \pm 4.9$ & $74.2 \pm 5.5$ & $76.5 \pm 5.2$\\
Koppula et al. \cite{koppula13IJRR} & $68.2 \pm 0.3$ & $71.1 \pm 1.9$ & $62.2 \pm 4.1$ & $66.4 \pm 2.6$ & $80.6 \pm 1.1$ & $81.8 \pm 2.2$ & $80.0 \pm 1.2$ & $80.9 \pm 1.6$\\
Koppula et al. \cite{koppula13icml} & $70.3 \pm 0.6$ & $74.8 \pm 1.6$ & $66.2 \pm 3.4$ & $70.2 \pm 2.2$ & $83.1 \pm 3.0$ & $87.0 \pm 3.6$ & $82.7 \pm 3.1$ & $84.8 \pm 3.3$ \\
Hu et al. \cite{Hu2014highlevel}\cite{Hu2014activity} & $70.0 \pm 0.3$ & $70.3 \pm 0.5$ & $67.8 \pm 0.2$ & $69.0 \pm 0.3$ & $79.0 \pm 6.2$ & $86.4 \pm 4.9$ & $78.8 \pm 5.9$ & $82.4 \pm 4.4$\\
Our Model (no latent)             & $67.1 \pm 0.4$ & $69.1 \pm 1.2$ & $65.6 \pm 1.5$ & $67.3 \pm 1.8$ & $79.0 \pm 2.0$ & $80.4 \pm 2.7$ & $78.5 \pm 2.0$ & $79.4 \pm 2.3$\\
Our Model (full)& $70.2 \pm 1.2$ & $71.1 \pm 1.8$ & $69.9 \pm 1.9$ & $70.5 \pm 1.9$ & $85.2 \pm 1.4$ & $90.3 \pm 1.9$ & $84.7 \pm 1.5$ & $87.4 \pm 1.7$\\
\bottomrule
\end{tabular}
\end{table*}

\begin{table*}
\renewcommand{\arraystretch}{1.3}
\caption{Test Performance on the \mbox{CAD-60} dataset with Uniform Segmentation. The standard error is also reported.}
\label{tab:results_cad60_hierarhical}
\centering
\begin{tabular}{lcccccccc}
\toprule
Methods & \multicolumn{4}{c}{Action} & \multicolumn{4}{c}{Environment} \\
\cmidrule(r){2-5}\cmidrule(r){6-9}
& Accuracy       & Precision      & Recall       & F-Score & Accuracy       & Precision      & Recall & F-Score   \\
\midrule
Single layer & - & - & - & - & $50.0\pm2.8$ & $63.0\pm2.5$ & $52.8\pm2.2$ & $57.5\pm2.3$\\
Hu et al. \cite{Hu2014highlevel}\cite{Hu2014activity} & $66.5 \pm 4.3$ & $71.1 \pm 2.6$ & $67.1 \pm 3.7$ & $67.7 \pm 3.4$ & $60.0 \pm 1.5$ & $71.0 \pm 2.5$ & $62.1 \pm 1.8$ & $63.0 \pm 2.1$\\
Our Model (full)& $74.4 \pm 4.0$ & $80.3 \pm 4.4$ & $81.0 \pm 1.6$ & $80.7 \pm 2.9$ & $60.6 \pm 0.5$ & $74.7 \pm 2.7$ & $62.5 \pm 1.0$ & $68.6 \pm 1.9$\\
\bottomrule
\end{tabular}
\end{table*}

In this section, we report the experimental results and compare the
performance of different models. Table~\ref{tab:results_cad120} shows
the performance of all the models during testing on the CAD-120
dataset. Both the performances of the action and activity recognition
are reported. For comparison, the results of both ground-truth
segmentation and motion-based segmentation are
reported. Table~\ref{tab:results_cad60_hierarhical} shows the
performance during testing on the CAD-60 dataset. 

Next, we analyze the results while referring to our research questions
posed in \secref{introduction}. 

\textbf{Importance of hierarchical model}. In
Table~\ref{tab:results_cad120}, \emph{Single Layer} refers to the
second baseline approach, wherein we learn a direct mapping from
video-level features to activity labels. There is no intermediate
layer of labels. The \emph{Single Layer} approach achieves an average
performance of over 70\% in both segmentation methods but with a large
standard error of approximately 5\%. In contrast, the other
hierarchical approaches outperform the \emph{Single Layer} approach by
at least 10 percentage points when using ground-truth segmentation and
5 percentage points when using motion-based segmentation. By
incorporating the layer of action labels, we can see significant
improvements in terms of recognizing activities. Therefore, temporal information, such as transitions between actions, is a very important aspect of activity recognition.

Table~\ref{tab:results_cad60_hierarhical} shows the results using the
CAD-60 dataset under similar experiment settings; however, the goal
here is to predict \emph{actions} together with the
\emph{environment}. We can see that the F-score of the environment
prediction is increased by over 11 percentage points when using the
hierarchical approach (\emph{full model}), which is significantly
better than the single-layer approach. The hierarchical approaches
also exhibit significant improvements in terms of precision and
recall. The increase in the mean is over 6 percentage points, with a reduced standard error rate.

\textbf{Importance of embedding the latent layer}.
To demonstrate the importance of using latent variables, we compare
the proposed model (\emph{full model}) to the model without augmenting
latent variables (\emph{no latent}). Table~\ref{tab:results_cad120}
shows that the \emph{full model} outperforms \emph{no latent} in terms of
recognition of both actions and activities. Notably, after adding the
latent variable, the precision and recall for activity is increased
by over $4$ and $5$ percentage points, respectively, using
ground-truth segmentation. When using motion-based segmentation, the
performance of \emph{full model} for an activity is increased by $10$
percentage points in terms of precision and $6$ percentage points in
terms of recall. The improvement is significant after using latent
variables. Note that the \emph{no latent} model is a special case of
the \emph{full model}, \ie, \emph{no latent} is equivalent to the
\emph{full model} when there is only one latent state. Here, we list
these models  separately to illustrate the effect of using multiple
latent states. In contrast, Table~\ref{tab:results_cad60_hierarhical}
only shows the performance of the \emph{full model} because the model
starts overfitting the data when more than one latent states are
applied to the model, \ie, \emph{no latent} (latent=1) achieves the
best performance. From this, we can see that the model is quite
flexible and that it can be used to fit data with varying levels of complexity by simply adjusting the number of latent states in the model.

\textbf{Importance of jointly modeling  activity and action}.
Hu et al. \cite{Hu2014highlevel}\cite{Hu2014activity} in
Table~\ref{tab:results_cad120} refers to a combination of the first
and third baseline approaches, where we used a two-step approach to
successively recognize actions and activities. This method shows
significant improvement over the \emph{Single layer}
approach. However, their approach is significantly outperformed by our
proposed hierarchical method (\emph{full model}) using both
segmentation methods. Notably, for activity recognition, the F-score
is increased by 3 percentage points using our proposed model, with an
increase of 4 percentage points in terms of precision and 6 percentage
points in terms of recall. For action recognition, the performance
gain in terms of F-score is approximately 3.7 percentage points and
includes significant improvements in both precision and recall. This
is because the \emph{full model} allows the interaction between the
low-level and high-level layers during both learning and inference,
and labels with the hierarchy are jointly estimated when making
predictions. Similar results were found using the \mbox{CAD-60}
dataset, see Table~\ref{tab:results_cad60_hierarhical}. We note that the performance is largely increased when using the \emph{full model}. The F-score is increased by 3 percentage points for predicting \emph{action} and \emph{environment} labels.

\textbf{Comparison with the state-of-the-art approaches}. The proposed
method was evaluated on both the \mbox{CAD-60} and
\mbox{\mbox{CAD-120}} datasets to provide a comparison with the state-of-the-art methods.

To be comparable with the other approaches, following
\cite{sung2012unstructured,koppula13IJRR}, we conduct similar
experiments on the CAD-60 dataset, where we group the actions based on
their environment labels and a separate model is learned and tested
for each of the groups. The results of these experiments are reported
in Table~\ref{tab:results_cad60}. We note that our model outperforms
\cite{sung2012unstructured} in all five environments. Compared with
the state of the art \cite{koppula13IJRR}, our model outperforms
\cite{sung2012unstructured} and \cite{koppula13IJRR} on most of the
\emph{environments}. On average, the precision of our model is the
same as in \cite{koppula13IJRR}, and the recall of the model
outperforms \cite{koppula13IJRR} by over $8$ percentage points,
achieving $80.8\%$ for precision and $80.1\%$ for recall. The average
F-Score is over $4\%$ percentage points better than in \cite{koppula13IJRR}.
\begin{table*}
\tabcolsep=0.11cm
\renewcommand{\arraystretch}{1.3}
\caption{Performance on the \mbox{CAD-60} dataset. Note that in these
  experiments, the actions are grouped based on their locations, and a
  separate model is trained and tested based the ENVIRONMENT label;
  therefore, the results are different from
  Table~\ref{tab:results_cad60_hierarhical}. The results are reported
  in terms of Precision (P. \%), Recall (R. \%), and F-Score
  (F. \%). The standard error of our model is reported.}
\label{tab:results_cad60}
\centering
\begin{tabular}{l*{18}{c}}
\toprule
Methods & \multicolumn{3}{c}{Bathroom} & \multicolumn{3}{c}{Bedroom} & \multicolumn{3}{c}{Kitchen} & \multicolumn{3}{c}{Living Room} & \multicolumn{3}{c}{Office} & \multicolumn{3}{c}{Average}\\
\cmidrule(r){2-4}\cmidrule(r){5-7}\cmidrule(r){8-10}\cmidrule(r){11-13}\cmidrule(r){14-16}\cmidrule(r){17-19}

 & P. \% & R. \% & F. \% & P. \% & R. \% & F. \% & P. \% & R. \% & F. \% & P. \% & R. \% & F. \% & P. \% & R. \% & F. \% & P. \% & R. \% & F. \%\\
\midrule
Sung et al. \cite{sung2012unstructured} & $72.7$ & $65.0$ & $68.9$ & $76.1$ & $59.2$ & $67.7$ & $64.4$ & $47.9$ & $64.7$ & $52.6$ & $45.7$& $49.2$ & $73.8$ & $59.8$ & $66.8$ & $67.9$ & $55.5$ & $61.7$\\
Koppula et al. \cite{koppula13IJRR} & $\textbf{88.9}$ & $61.1$ & $75.0$ & $73.0$ & $66.7$ & $69.9$ & $\textbf{96.4}$ & $85.4$ & $\textbf{90.9}$ & $69.2$ & $68.7$ & $69.0$ & $76.7$ & $75.0$ & $75.9$ & $\textbf{80.8}$ & $71.4$ & $76.1$\\
Our Model & $77.6$ & $\textbf{81.5}$ & $\textbf{78.1}$ & $\textbf{81.8}$ & $\textbf{76.9}$ & $\textbf{78.8}$ & $88.2$ & $\textbf{92.0}$ & $90.1$ & $\textbf{80.6}$ & $\textbf{75.9}$ & $\textbf{78.3}$ & $\textbf{81.7}$ & $\textbf{75.1}$ & $\textbf{78.4}$ & $\textbf{80.8}$ & $\textbf{80.1}$ & $\textbf{80.5}$ \\
Our Model (std. err.) & $(6.9)$ & $(5.7)$ & $(6.3)$ & $(2.7)$ & $(3.5)$ & $(3.1)$ & $(5.2)$ & $(2.6)$ & $(3.9)$ & $(6.6)$ & $(7.3)$ & $(7.0)$ & $(5.4)$ & $(6.2)$ & $(5.8)$ & $(5.4)$ & $(5.1)$ & $(5.3)$\\
\bottomrule
\end{tabular}
\end{table*}

Table~\ref{tab:results_cad120} compares the performance of different
approaches on the \mbox{\mbox{CAD-120}} datasets. Similar to
\cite{Hu2014highlevel}, Koppula et al. \cite{koppula13icml} use a
two-step approach to infer high-level activity labels only after the
actions are estimated. Benefiting from the joint estimation of action
and activity, our \emph{full model} outperforms the state-of-the-art
models in terms of both action and activity recognition
tasks. Notably, using ground-truth segmentation, the F-score is
improved by approximately 4 percentage points for recognizing
actions. Based on motion segmentation, the activity recognition
performance is improved by over 2 percentage points in terms of F-Score. 

\begin{figure*}
\centering
\subfloat[]{\includegraphics[width=0.4\linewidth]{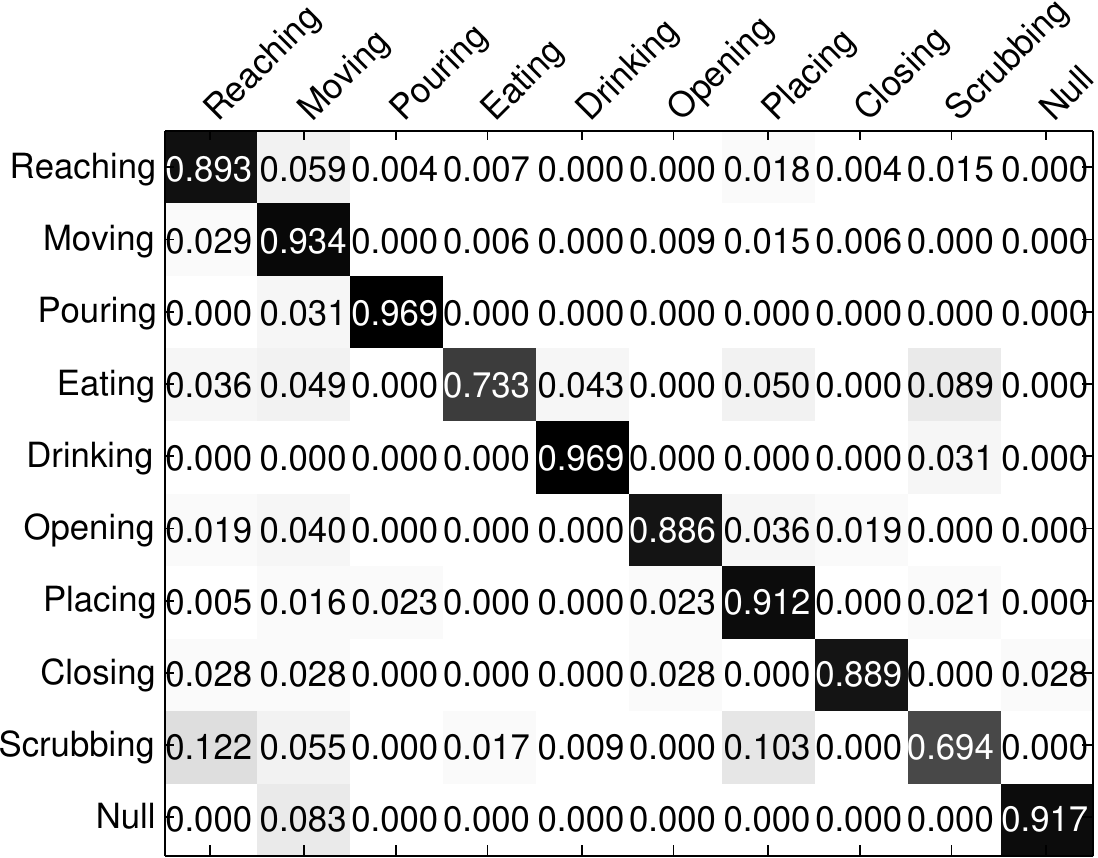}%
\label{fig:confmat_actions}}%
\subfloat[]{\includegraphics[width=0.45\linewidth]{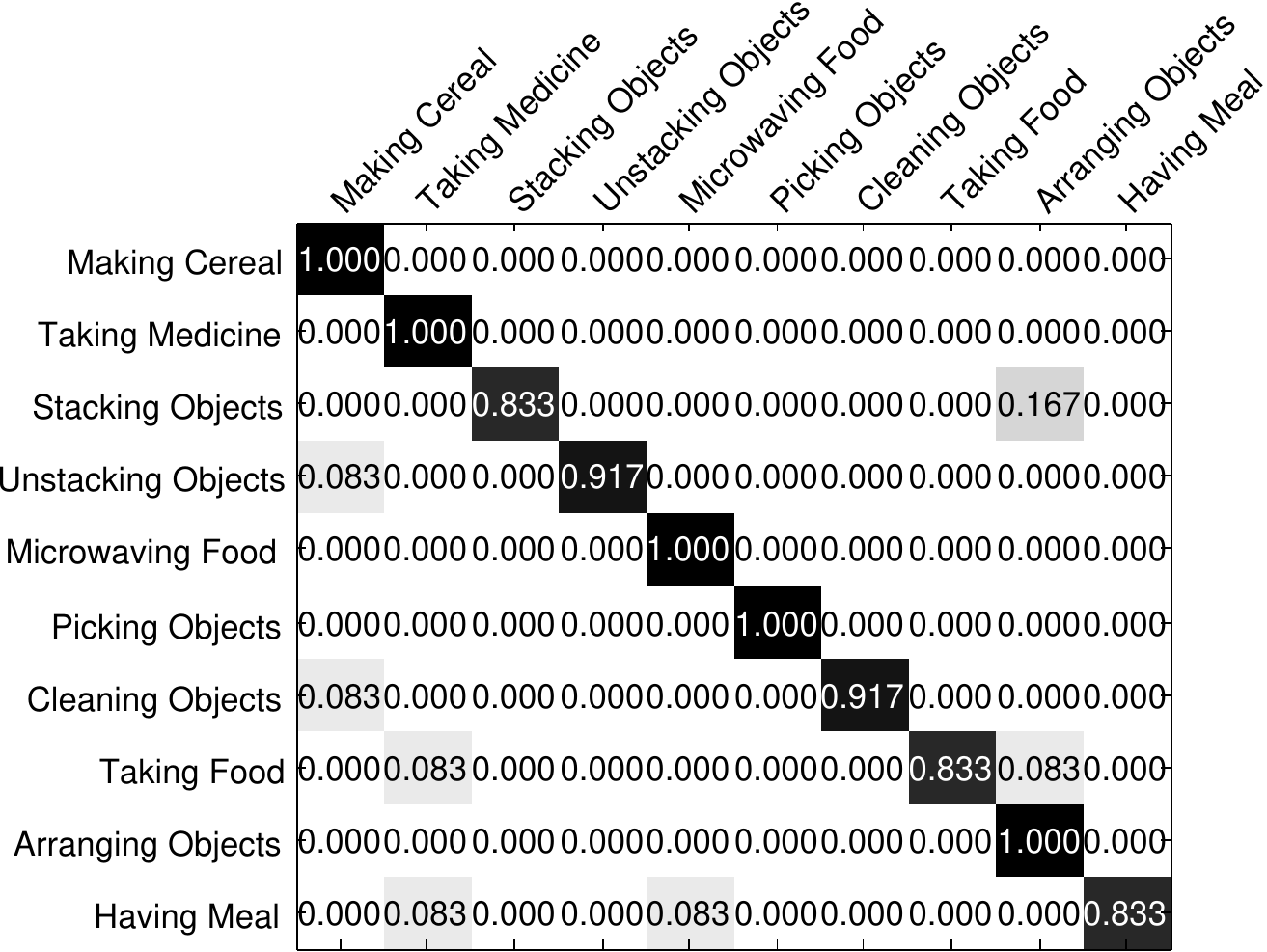}%
\label{fig:confmat_activity}}%
\caption{Confusion matrix over different \emph{action} \protect\subref{fig:confmat_actions} and \emph{activity} \protect\subref{fig:confmat_activity} classes. Rows are ground-truth labels, and columns are the detections.}
\label{fig:confmat}
\end{figure*}

\fig{confmat} shows the confusion matrix of both the action and
activity classification results. The most difficult action class is
\emph{scrubbing}. This task is sometimes confused with \emph{reaching}
and \emph{placing}. The overall performance of the activity
recognition is very good, with most of the activities being correctly
classified. The more difficult case is to distinguish between
``stacking objects'' and ``arranging objects''. Overall, we can see
that high values are found on the diagonal using both segmentation
methods, which demonstrates the good performance of our system.

%

%
\section{Conclusion}

In this paper, we present a hierarchical approach that simultaneously
recognizes actions and activities based on \mbox{RGB-D} data. The
interactions between actions and activities are captured by a
\mbox{Hidden-state} CRF framework. In this framework, we use the
latent variables to exploit the underlying structures of actions. The
prediction is based on the joint interaction between activities and
actions, which is in contrast to the traditional approach, which only
focuses on one of them. Our results show a significant improvement
when using the hierarchical model compared to using the single-layered approach. The results also demonstrate the effectiveness of adding a latent layer to the model and the importance of jointly estimating actions and activities. Finally, we show that the proposed hierarchical approach outperforms the state-of-the-art methods on two benchmark datasets.


\bibliographystyle{IEEEtran}
\bibliography{IEEEabrv,refs}


\ifCLASSOPTIONcaptionsoff
  \newpage
\fi

\begin{IEEEbiography}[{\includegraphics[width=1in,height=1.25in,clip,keepaspectratio]{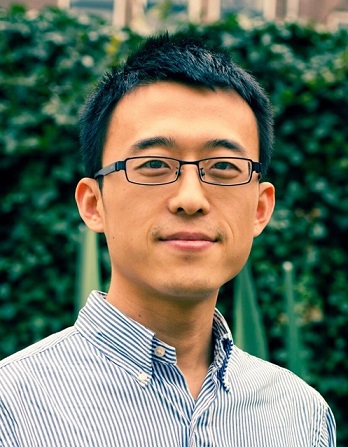}}]{Ninghang Hu}
is currently a Ph.D. candidate in the Informatics Institute at the University of Amsterdam, The Netherlands. His research interests are in machine learning and robot vision, with a focus on human activity recognition and data fusion. He received his M.Sc. in Artificial Intelligence from the same university in 2011 and his B.Sc. in Software Engineering from Xidian University, China in 2008. In 2011, he was a trainee at TNO (the Netherlands Organisation for Applied Scientific Research). 
\end{IEEEbiography}

\begin{IEEEbiography}
[{\includegraphics[width=1in,height=1.25in,clip,keepaspectratio]{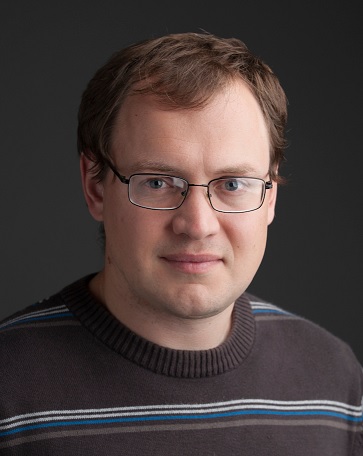}}]
{Gwenn Englebienne}
received the Ph.D. degree in computer science from the University of Manchester in 2009. He has since focused on automated analysis of human behavior at the University of Amsterdam, where he has developed computer vision techniques for tracking humans across large camera networks and machine learning techniques to model human behavior from networks of simple sensors. His main research interests are in models of human behavior, especially their interaction with other humans, with the environment, and with intelligent systems.
\end{IEEEbiography}


\begin{IEEEbiography}
[{\includegraphics[width=1in,height=1.25in,clip,keepaspectratio]{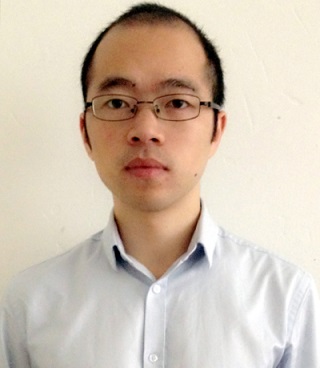}}]
{Zhongyu Lou}
received his B.Sc. and M.Sc. degrees from Xidian University, Xi'An,
China, in 2008 and 2011, respectively. He is currently pursuing his
Ph.D. degree at the Intelligent Systems Lab Amsterdam, University of Amsterdam, The Netherlands. His main research interests include computer vision, image processing and machine learning.
\end{IEEEbiography}

\begin{IEEEbiography}
[{\includegraphics[width=1in,height=1.25in,clip,keepaspectratio]{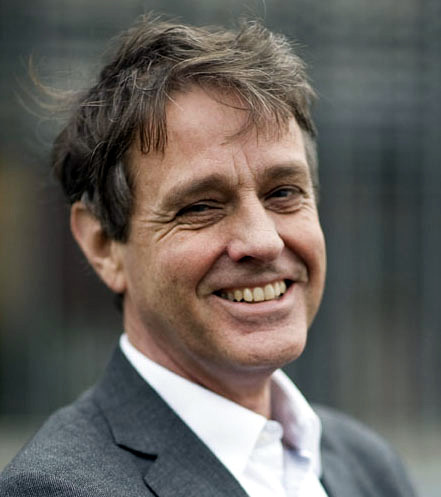}}]
{Ben Kr\"{o}se}
is a Professor of ``Ambient Robotics'' at the University of Amsterdam
and a Professor of ``Digital Life'' at the Amsterdam University of
Applied Science. His research focuses on robotics and interactive
smart devices, which are expected to be widely applied to smart
services to ensure the health, safety, well-being, security and
comfort of users.  In the fields of artificial intelligence and
autonomous systems, he has published 35 papers in scientific journals,
edited 5 books and special issues, and submitted more than 100
conference papers. He owns a patent on multi-camera surveillance. He
is a member of the IEEE, Dutch Pattern Recognition Association and Dutch AI Association.
\end{IEEEbiography}




\end{document}